\documentclass[letterpaper]{article} 
\usepackage{aaai2026}  
\usepackage{times}  
\usepackage{helvet}  
\usepackage{courier}  
\usepackage[hyphens]{url}  
\usepackage{graphicx} 
\usepackage{natbib}  
\usepackage{caption} 
\makeatletter
\newcommand{\thickhline}{%
    \noalign {\ifnum 0=`}\fi \hrule height 1pt
    \futurelet \reserved@a \@xhline
}
\makeatother
\usepackage{multirow}
\usepackage{algorithm}
\usepackage{algorithmic}
 \usepackage{amsmath}

\usepackage{colortbl} 
\usepackage{xcolor}
\usepackage[utf8]{inputenc}
\usepackage{url}
\usepackage{booktabs}
\usepackage{amssymb}
\usepackage{bbding}
\usepackage{pifont}
\usepackage{wasysym}
\usepackage{utfsym}
\usepackage{fontawesome}
\usepackage{newfloat}
\usepackage{listings}
\DeclareCaptionStyle{ruled}{labelfont=normalfont,labelsep=colon,strut=off} 
\floatstyle{ruled}
\newfloat{listing}{tb}{lst}{}
\floatname{listing}{Listing}
\title{Splats in Splats: Robust and Effective 3D Steganography towards \\ Gaussian Splatting}
\author{
      Yijia Guo \textsuperscript{\rm 1}\equalcontrib, 
      Wenkai Huang \textsuperscript{\rm 2,3}\equalcontrib, 
      Yang Li \textsuperscript{\rm 1}, 
      Gaolei Li \textsuperscript{\rm 2,3}\thanks{Corresponding authors.}, 
      Hang Zhang \textsuperscript{\rm 5}\\
      Liwen Hu \textsuperscript{\rm 1}, 
      Jianhua Li \textsuperscript{\rm 2,3}, 
      Tiejun Huang \textsuperscript{\rm 1}, 
      Lei Ma \textsuperscript{\rm 1,4}\footnotemark[2]\\
}
\affiliations{
    \textsuperscript{\rm 1}National Key Laboratory for Multimedia Information Processing, Peking University\\
    \textsuperscript{\rm 2}School of Computer Science, Shanghai Jiao Tong University\\
    \textsuperscript{\rm 3}Shanghai Key Laboratory of Integrated Administration Technologies for Information Security, Shanghai Jiao Tong University\\
    \textsuperscript{\rm 4}National Biomedical Imaging Center, Peking University \quad
    \textsuperscript{\rm 5}Cornell University \\
    \{2301112015, ly0376, liwenhu, tjhuang, leima\}@stu.pku.edu.cn, \{sjtuhwk, gaolei\_li, lijh888\}@sjtu.edu.cn, hz459@cornell.edu

}

\usepackage{bibentry}

\begin{document}

\maketitle

\begin{abstract}
3D Gaussian splatting (3DGS) has demonstrated impressive 3D reconstruction performance with explicit scene representations. Given the widespread application of 3DGS in 3D reconstruction and generation tasks, there is an urgent need to protect the copyright of 3DGS assets. However, existing copyright protection techniques for 3DGS overlook the usability of 3D assets, posing challenges for practical deployment.
Here we describe \textbf{\textit{splats in splats}}, the first 3DGS steganography framework that embeds 3D content in 3DGS itself without modifying any attributes. To achieve this, we take a deep insight into spherical harmonics (SH) and devise an importance-graded SH coefficient encryption strategy to embed the hidden SH coefficients. Furthermore, we employ a convolutional autoencoder to establish a mapping between the original Gaussian primitives' opacity and the hidden Gaussian primitives' opacity. Extensive experiments indicate that our method significantly outperforms existing 3D steganography techniques, with \textbf{5.31\%} higher scene fidelity and \textbf{3\textbf{$\times$}} faster rendering speed, while ensuring security, robustness, and user experience. 

\end{abstract}    

\section{Introduction}
\label{sec:intro}
Building on the success of utilizing a discrete 3D Gaussian representation for scenes, 3D Gaussian Splatting (3DGS) \cite{3dgs} significantly accelerates the training and rendering speed of radiance fields with explicit scene representations. Given the widespread application of 3DGS in 3D reconstruction and generation tasks \cite{yi2023gaussiandreamer, yi2024gaussiandreamerpro}, the identification and attribution of 3DGS assets are crucial to ensure their secure usage and protect against copyright infringement. Steganography can play a key role in verifying the provenance of 3DGS assets and preventing both accidental and deliberate misuse. In addition to ensuring security, fidelity, and robustness as observed in traditional steganography, the steganography technique for 3DGS must fulfill the following two criteria:

\textbf{a) Protecting the 3DGS asset itself rather than the rendered image.}
A straightforward approach for 3DGS steganography is embedding specific information directly into the rendered images from the 3DGS \cite{huang2024gaussianmarker}. However, it solely safeguards the copyright of the rendered images from specific viewpoints instead of the copyright of the 3DGS itself. Malicious users may generate new samples employing different rendering strategies after stealing the core model, circumventing the external steganography anticipated by the creators.

\begin{figure*}
  \centering
  \includegraphics[width=0.91\linewidth]{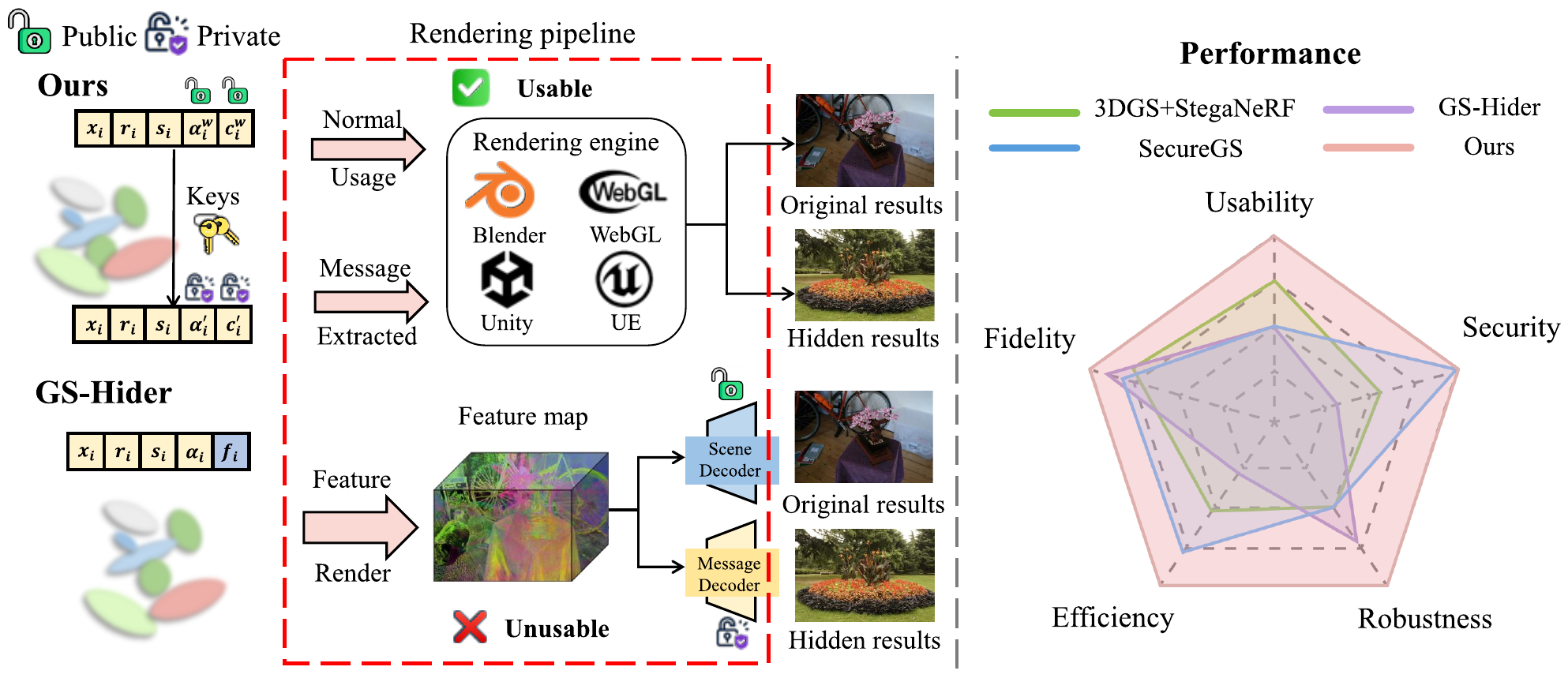}
  \captionof{figure}{\textbf{Left:} GS-Hider and our method's rendering pipeline. GS-Hider \cite{zhang2024gshider} employs a coupled feature field and neural decoders to render the original and hidden scenes simultaneously, affecting user’s standard utilization. We retain the vanilla 3DGS pipeline to preserve user experience. \textbf{Right:} Comparison of different 3DGS steganography methods. Existing works all have shortcomings in terms of robustness, fidelity, efficiency, and usability. Our method can maximize the fidelity of the original scene while ensuring the rendering speed and usability of 3DGS, as well as the security of the steganography. }
  \label{fig1}
\end{figure*}

\textbf{b) Ensuring the usability of 3DGS assets.} Steganography techniques must not disrupt normal usage of 3DGS assets. Digital steganography for 2D images \cite{baluja2019hiding, xu2022robust}, videos \cite{mou2023large, luo2023dvmark} and Neural Radiance Fields (NeRF) \cite{li2023steganerf,jang2024waterf} have all ensured this aspect. However, existing 3DGS steganography techniques such as GS-Hider \cite{zhang2024gshider} and SecureGS \cite{zhang2025securegs} modify the attributes and rendering pipeline of vanilla 3DGS, as shown in Fig.~\ref{fig1}. These methods greatly affect users’ standard utilization since the modified 3DGS asset poses challenges for practical deployment in vanilla 3DGS rendering engines and downstream tasks. The fundamental solution is to keep the attributes of vanilla 3DGS.
\textbf{Is there a solution that can embed 
 hidden information in 3DGS itself without modifying any attributes of the vanilla 3DGS?}

To fulfill the above demands, we propose an effective, flexible and robust steganography framework named \textit{\textbf{splats in splats}}. To the best of our knowledge, splats in splats is the first framework that embeds 3D content into the vanilla 3DGS while fully preserving its attributes. To achieve this, we take a deep insight into spherical harmonics (SH) and devise an importance-graded SH coefficient encryption/decryption strategy to embed the hidden SH coefficients. Furthermore, we employ a convolutional autoencoder to establish a mapping between the original Gaussian primitives' opacity and the hidden Gaussian primitives' opacity. Splats in splats can also embed other types of digital content, like images. Our main contributions are as follows:
\begin{itemize}
\item[$\bullet$] 
 We propose splats in splats, the first steganography framework that embeds 3D content in 3DGS itself without modifying any attributes of the vanilla 3DGS. 
\item[$\bullet$]
Building on a deep insight into the Gaussian primitives' attributes, we devise importance-graded SH coefficient encryption and autoencoder-assisted opacity mapping to preserve the structural integrity of the vanilla 3DGS.

\item[$\bullet$] 
Extensive experiments indicate that splats in splats achieves state-of-the-art fidelity and efficiency while ensuring security, robustness, and user experience.
\end{itemize}

\section{Related Works}
\label{sec:formatting}
\subsection{3D Gaussian Splatting}
3D Gaussian splatting \cite{3dgs} significantly accelerates the training and rendering speed of radiance fields. This breakthrough immediately attracted countless attention. Recently,  many researchers have focused on improving the rendering quality \cite{yu2024mip}, effectiveness \cite{fan2024instantsplat,chen2024mvsplat} and the storage \cite{navaneet2024compgs} of 3DGS, extending it to dynamic 3D scenes \cite{wu20234d,li2024st}, large-scale outdoor scenes \cite{liu2024citygaussian}, and high-speed scenes \cite{xiong2024event3dgs,yu2024evagaussians,guo2024spikegs}, relaxing its restrictions on camera poses \cite{muller2022instant} and sharp images \cite{chen2024deblur}. 3DGS is also widely used in the area of inverse rendering \cite{guo2024prtgs,liang2023gsir,gao2023relightable}, 3D generation \cite{yi2023gaussiandreamer} and 3D editing \cite{chen2023gaussianeditor}. Patently, 3DGS is becoming a mainstream 3D asset. Thus, the management of the copyright of 3DGS has been emerging as an urgent problem.

\begin{figure*}[t]
  \centering
  \includegraphics[width=0.92\linewidth]{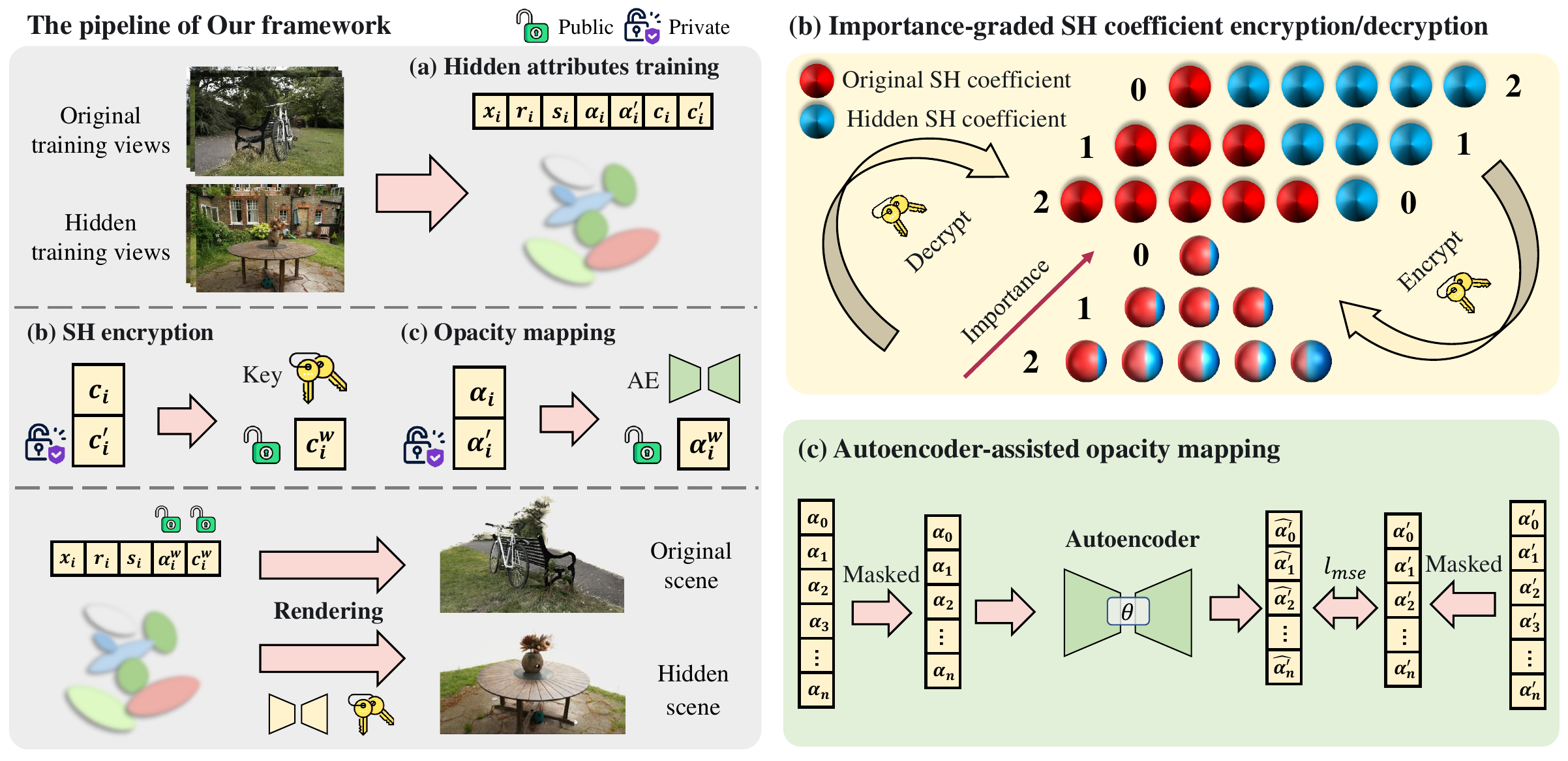}
  \caption{\textbf{Overview of the Framework}. To ensure a seamless user experience, we preserves the same attributes as the vanilla 3DGS, while enabling the extraction of embedded 3D content for owners. The information embedding and extraction process involves three key steps: \textbf{a) Hidden attributes training}: We utilize the original and hidden views to train two sets of SH coefficients and opacity, while ensuring that both sets share the same Gaussian primitive locations.
  \textbf{b) Importance-graded SH coefficient encryption/decryption}: We prioritize SH coefficients by significance, embedding more important hidden SH coefficients into higher-order components of the original SH via bit-shifting, aiming to achieve superior rendering fidelity and high robustness against noise attack.
  \textbf{c) Autoencoder-assisted opacity mapping}: We apply a threshold to discard trivial hidden opacities, subsequently employing an convolutional autoencoder to model the transformation from original to hidden opacity.}
  \label{fig2}
\end{figure*}

\subsection{3D Steganography}
3D steganography techniques place greater emphasis on the unique characteristics of 3D representations rather than 2D images \cite{baluja2019hiding, xu2022robust}, videos \cite{mou2023large, luo2023dvmark} or large-scale generative models \cite{fernandez2023stable, wen2024tree}. Existing 3D steganography techniques primarily focus on meshes or point clouds, employing domain transformations \cite{li2023zero, kallel20233d} or homomorphic encryption \cite{van20233d} for information hiding. 

Recently, steganography techniques on implicit 3D representations like NeRF have gained prominence. StegaNeRF \cite{li2023steganerf} embedding secret images into 3D scenes by fine-tuning NeRF's weights  while preserving the original visual quality. WaterRF \cite{jang2024waterf} embeded binary messages using discrete wavelet transform, achieving great performance. When considering 3DGS, GS-Hider \cite{zhang2024gshider} utilized a scene and a message decoder to disentangle the hidden message from the original scene. However, GS-Hider not only caused a degradation in both rendering speed and visual quality but also damaged the vanilla 3DGS architecture, resulting in limited effectiveness in practical deployment. SecureGS, which is built on Scaffold-GS \cite{scaffoldgs}, faces the same problems. So there is an urgent need for 3DGS steganography research that ensures copyright protection while maintaining the efficiency and usability of the vanilla 3DGS framework.


\section{Preliminary}
In 3DGS, every Gaussian primitive $G$ is defined by a full 3D covariance matrix $\Sigma$ in world space centered at $x \in \mathbb{R}^3$:
\begin{align}
    G(x)=e^{-1/2(x)^T\Sigma^{-1}(x)}. \
\end{align}
The covariance matrix $\Sigma$ of a 3D Gaussian primitive can be described as a rotation matrix \(R\) and a scale matrix \(S\) and independently optimize of both them. 
\begin{align}
    \Sigma=RSS^TR^T.
\end{align}
 Further, we utilize the method of splatting to project our 3D Gaussian primitives to 2D camera planes for rendering:
\begin{align}
    \Sigma'=JW\Sigma W^TJ^T.
\end{align}
Where \(J\) is the Jacobian of the affine approximation of the projective transformation and \(W\) is the viewing transformation.
Following this, the pixel color is obtained by alpha-blending \(N\) sequentially layered 2D Gaussian splats:
\begin{align}
    C=\sum_{\substack{i\in N}}c_{i}\alpha_{i}\prod_{j=1}^{i-1}(1-\alpha_{j}). \label{point rendering}
\end{align}
Where $c_{i}$ is the color of each point and $\alpha_{i}$
is given by evaluating a 2D Gaussian with covariance $\Sigma$ multiplied with a learned per-point opacity.

In summary, each 3D Gaussian primitive is defined by five attributes: \(\{x_i, r_i, s_i, \alpha_i, c_i\}\). Specifically, \(x_i\) and \(s_i\) are represented as \(1 \times 3\) vectors, while \(r_i\) is formatted as a \(1 \times 4\) vector, and \(\alpha_i\) is a scalar value. Notably, \(c_i\) is constituted as a \(n \times 3\) matrix of Spherical Harmonic (SH) coefficients, which effectively compensates for view-dependent effects.

\section{Methodology}
\label{sec:method}

\subsection{Overview}
The overall workflow of our method is depicted in Fig.~\ref{fig2}. Our objective is to seamlessly embed a 3D content into 3DGS assets while ensuring that the typical usage pipeline for regular users remains unaffected. Accordingly, the asset owner can leverage a private key to recover the hidden attributes and extract the 3D content, enabling robust verification. To achieve this, we first pre-train hidden SH coefficients and opacity for each 3D Gaussian primitive to align with the hidden scene. Subsequently, we devise importance-graded SH coefficient encryption/decryption and autoencoder-assisted opacity mapping strategies to effectively accomplish 3D information embedding and extraction. With these strategies, we can not only protect the copyright of 3DGS assets but also maintain the essential attributes of vanilla 3DGS, thereby ensuring its usability.

\subsection{An Insight in Spherical Harmonics} \label{insight}
\begin{figure}[t]
  \centering
  \includegraphics[width=1\linewidth]{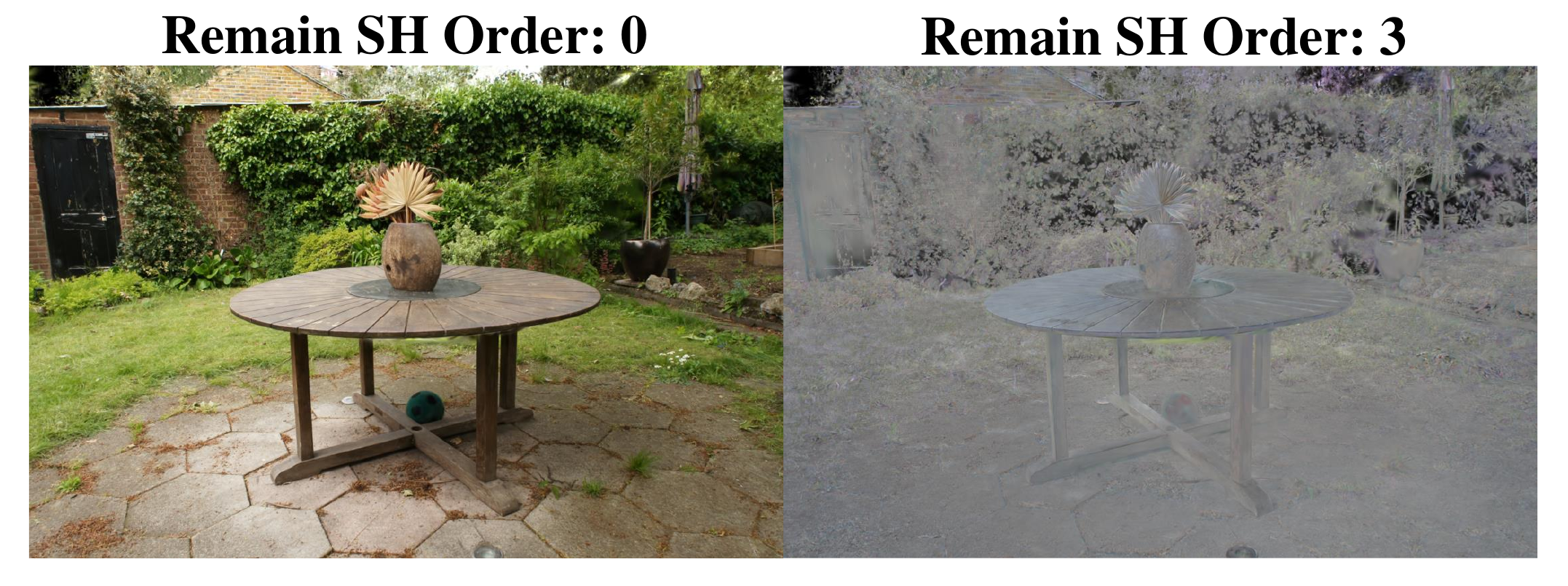}
  \caption{Importance of Spherical Harmonics order. We retain the specified order of SH and render the image while setting other orders to zero. Higher-order SH contains only a small amount of High-frequency information. }
  \label{fig_SH}
\end{figure}

Any function $F(s)$ is defined on the sphere $S$ can be represented as a set of SH basis functions:
\begin{align}
    F(s)\approx\sum_{l=0}^{q-1}\sum_{m=-l}^{l}f_l^mY_l^m(s),\label{sh}
\end{align}
and the basis functions are defined as:
\begin{align}
   Y^m_l(\theta,\phi)=K^m_le^{im\phi}P_l^{|m|},l \in N,m \in (-l,l),
\end{align}
where $P_l^{|m|}$ are the associated Legendre polynomials and $K^m_l$ are the normalization constants:
\begin{align}
    K^m_l=\sqrt{\frac{(2l+1)(l-|m|)!}{4\pi (l+|m|)!}}.
\end{align}
Low values of $l$ (called the band index) represent low-frequency 
basis functions over the sphere while high values of $l$ represent high-frequency 
basis functions.
In most cases, higher-frequency reflections occupy only a small proportion of the scene. Consequently, the contribution of higher-order spherical harmonic coefficient is minimal, leading to information redundancy in these higher-order terms, as illustrated in Fig.~\ref{fig_SH}. Embedding information within these terms would pose considerable challenges for detection and would ultimately preserve fidelity. In this paper, we capitalize on this property of spherical harmonics to embed information while simultaneously improving resilience against noise attacks that target spherical harmonics.

\subsection{3D Information Embedding}
As illustrated in Fig.~\ref{fig2}, we embed the trained hidden SH coefficients and opacities into the original Gaussian primitives. For the SH coefficient encryption, we prioritize SH coefficients by order, embedding the more significant hidden coefficients into the higher-order components of the original SH through bit shifting. For the opacity mapping, we set a threshold to filter out insignificant hidden opacities and utilize an convolutional autoencoder to learn the mapping from the original opacity to the hidden opacity. 

\subsubsection{Importance-graded SH Coefficient Encryption}
For the original SH coefficients \( c_i \in \mathbb{R}^{n \times 3} \) and the hidden SH coefficients \( c_i' \in \mathbb{R}^{n \times 3} \), let \( c_{i,j} \) denote the \( j \)-th \((0 \leq j \leq n-1) \) component of \( c_i \). Then, the order of \( c_{i,j} \) is  \( \lfloor \sqrt{j} \rfloor\) according to the definition, where \( \lfloor \cdot \rfloor \) denotes the floor function. The lower the order of \( c_{i,j} \), the more significant the coefficient is in the Gaussian primitive. Drawing from the insight from Sec.~\ref{insight}, we embed more bits of the low-order \( c_{i,j}' \) into the higher-order \( c_{i,j} \) to achieve higher quality. For computational convenience, we scale these coefficients and represent them in integer form as binary values of \(\{0, 1\}^\gamma\).

Firstly, we nullify the lower bits of \( c_{i,j} \) \textbf{based on its graded importance:}
\begin{equation}
\tilde{c}_{i,j} = c_{i,j} \enspace \& \sim((1 << (k + \lfloor \sqrt{j} \rfloor)) - 1).
\label{eq：SH}
\end{equation}
Where \(<<\) denotes left bit shifting operation, \(\sim\) operator represents the bitwise NOT operation, and \( k \) represents the bit shifting length corresponding to the 0-order coefficient. 
Subsequently, we shift \({c}_{i,j}'\) to the corresponding position and \textbf{perform an exclusive OR operation with \(\tilde{c}_{i,j}\):}
\begin{equation}
{c}_{i,j}^{w} = \tilde{c}_{i,j} \oplus ({c}_{i,n-1-j}' >> (\gamma - (k + \lfloor \sqrt{j} \rfloor))).
\end{equation}
Where \(>>\) denotes right bit shifting operation, \( \gamma \) represents the maximum bit length, and \( {c}_{i,j}^{w} \) denotes the fused SH coefficient. The \( n-1-j \) term indicates that \({c}_{i,j}'\) has been reversed to match our importance-graded selection. Building on this strategy, we maintain the inherent attributes of original 3DGS, while achieving superior rendering fidelity.

\subsubsection{Autoencoder-assisted Opacity Mapping}
To balance the quality of both the original and hidden scenes, we also incorporate hidden opacity attributes and employ a convolutional autoencoder to learn the mapping from original opacity to hidden opacity. Let the original opacity be denoted as \(\alpha \in \mathbb{R}^{N \times 1}\) and the hidden opacity attribute as \(\alpha' \in \mathbb{R}^{N \times 1}\). Here $N$ is the num of Gaussian primitives after training. We first set a threshold \(\tau\) to obtain the indices of the more significant hidden opacity values: 
\begin{equation}
\mathcal{I} = \{ i \mid \alpha'_i > \tau, \; i \in \{1, 2, \dots, N\} \}.
\label{eq：op}
\end{equation}
Notably, we store the coordinates \( x_{\mathcal{I}} \) of the Gaussian primitives at the indices \( \mathcal{I} \), which are then used in the hidden opacity estimation process. Then we denote \(\alpha_{\mathcal{I}}\) and \(\alpha_{\mathcal{I}}'\) as the values of \(\alpha\) and \(\alpha'\) at \(\mathcal{I}\), respectively. 
Based on the observation that \( \alpha_i \) and \( \alpha'_i \) exhibit complementary relationships at many positions, we apply an autoencoder to \(1 - \alpha_{\mathcal{I}}\), performing encoding and decoding to obtain \(\hat{\alpha}_{\mathcal{I}}'\). We then utilize the Mean Squared Error (MSE) to train the autoencoder, learning the mapping from \(\alpha_{\mathcal{I}}\) to \(\alpha_{\mathcal{I}}'\):
\begin{equation}
W_p^* = \arg\min_{\mathcal{E}, \mathcal{D}} \ell_{mse} (\mathcal{D}(\mathcal{E}(1 - \alpha_{\mathcal{I}})), \alpha_{\mathcal{I}}').
\end{equation}
Here, \(\mathcal{E}\) and \(\mathcal{D} \) represent the encoder and decoder, respectively. To ensure real-time rendering, our autoencoder is composed of several simple convolutional and deconvolutional layers. The trained model parameters \( W_p^*\) then serve as the private attribute for the owner. Through this strategy, we reduce the storage requirements for the asset owner, while preserving the usability of the modified 3DGS.

\begin{figure*}[t]
  \centering
  \includegraphics[width=0.85\linewidth]{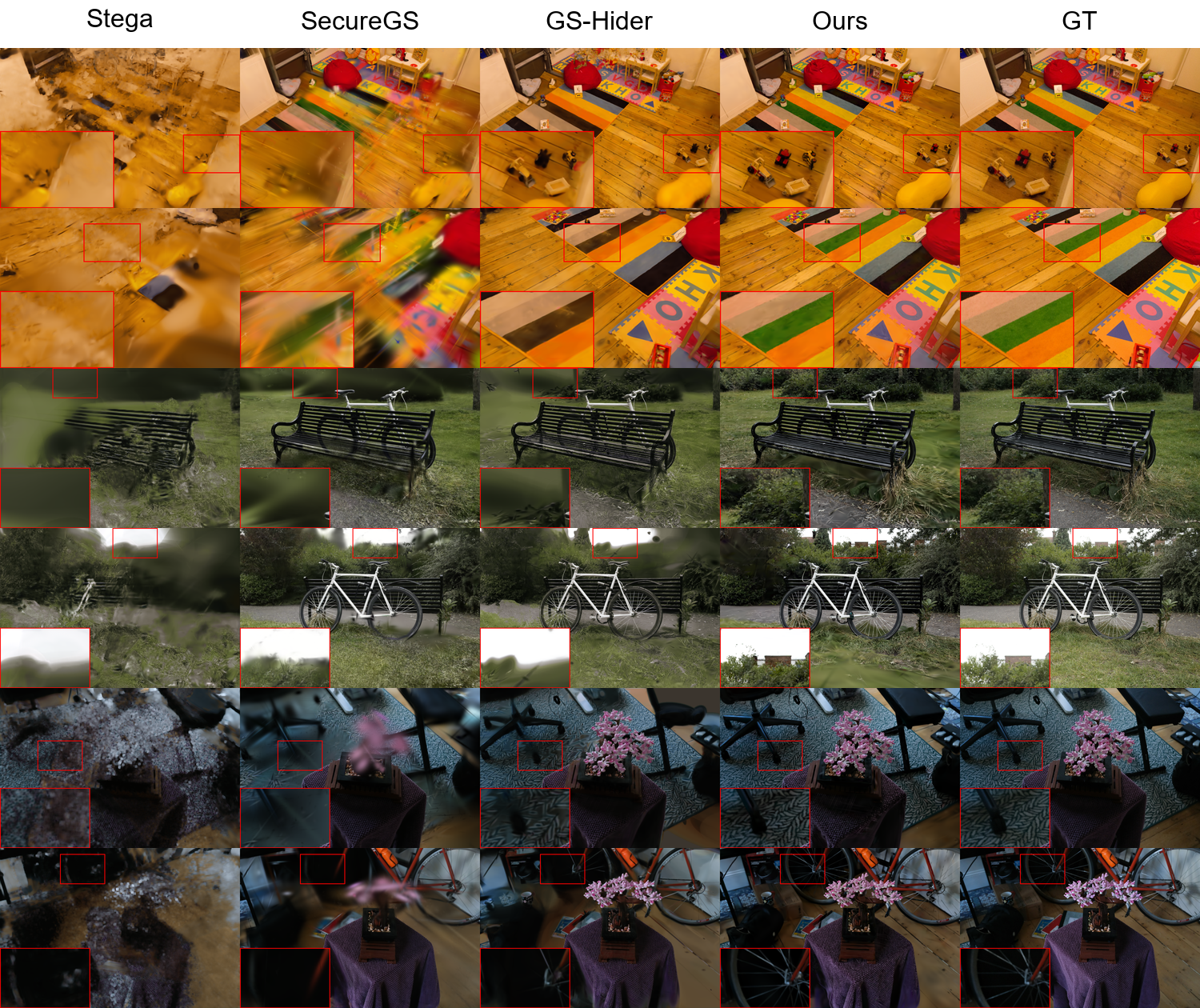}
  \caption{Qualitative comparisons on Mipnerf360 datasets. Our method achieves sharper results with more consistent structures.}
  \label{fig_compare}
\end{figure*}

\subsection{3D Information Extraction}
In cases of suspected copyright infringement, we can use SH coefficient decryption and opacity estimation to recover the hidden \({c}_{i,j}'\) and \(\alpha'_i \):
\begin{equation}
{c}_{i,j}' = {c}_{i,n-1-j}^{w} \enspace \& \enspace (1 << (k + \lfloor \sqrt{n-1-j} \rfloor)),
\end{equation}
\begin{equation}
     \alpha'_{\mathcal{I}}  =  \mathcal{D}_p(\mathcal{E}_p(1 - \alpha_{\mathcal{I}})).
\end{equation}
Where \( {c}_{i,j}^{w} \) is the public SH coefficient, and \(\mathcal{E}_p\) and \(\mathcal{D}_p\) represent the encoder and decoder obtained from \( W_p^* \), respectively. In practical scenarios, we can choose to retain only the recovered low-order SH coefficients to defend against potential noise attacks. For indices outside of the set \(\mathcal{I}\) (i.e., positions from 1 to \(N\) not included in \(\mathcal{I}\)), we set the hidden opacity to zero, as these locations are considered less significant. Finally, we use the recovered attributes \(\{x_i, r_i, s_i, \alpha_i', c_i'\}\) as the hidden 3DGS, which contains the embedded 3D information.


\begin{table*}[t]
\setlength{\tabcolsep}{0.9mm}
  \centering

    \resizebox{0.95\linewidth}{!}
    {

    \begin{tabular}{p{2.6cm}|c|c|c|c|c|c|c|c|c|c|c}
     \hline\thickhline
\rowcolor{black!10}  \centering Method&Type&Bicycle& Flowers& Garden& Stump &Treehill& Room &Counter &Kitchen &Bonsai& Average\\
\hline
\multirow{2}{*}{\parbox{2.6cm}{\centering  3DGS+ \\StegaNeRF }} &Scene &22.789& 19.811& 23.140 &24.260 &21.465 &28.659& 26.400& 25.039 &25.521& 24.120\\
 &Message&15.990& 14.753 &16.615 &16.513 &17.543 &17.213& 16.706 &17.534 &17.259& 16.681\\
 \hline
  \multirow{2}{*}{\parbox{2.6cm}{\centering GS-Hider }}&Scene&24.018&20.109& \underline{26.753} &24.573 &21.503 &28.865 &27.445 &29.447&29.643& 25.817\\
 ~ &Message&\underline{28.219}& \underline{26.389}& \textbf{32.348}& \underline{25.161}& 20.276 &\underline{22.885} &20.792 &\underline{26.690} &\underline{23.846}& \underline{25.179}\\
 \hline
 

  \multirow{2}{*}{\parbox{2.6cm}{\centering SecureGS  }}&Scene&\underline{24.237}& \underline{20.566}& 26.624 &\textbf{25.523} &\textbf{22.218} &\textbf{31.223} &\underline{28.499} &\underline{29.779}& \underline{30.499}& \underline{26.574}\\
 ~ &Message&21.798&23.838& 26.883&24.615&\textbf{23.778} &21.576 &20.851 &23.240 &21.367& 23.679\\
 \hline

   \multirow{2}{*}{\parbox{2.6cm}{\centering Ours}}&Scene &\textbf{24.431}& \textbf{20.685}& \textbf{26.831}&\underline{25.414} &\underline{22.156} &\underline{30.930} &\textbf{28.642} &\textbf{30.699}& \textbf{30.953}& \textbf{26.749}\\
  
                    ~ &Message &\textbf{28.988}& \textbf{29.006}& \underline{29.013}& \textbf{28.811}& \underline{22.794} &\textbf{23.355} &\textbf{22.514} &\textbf{28.345} &\textbf{25.826}& \textbf{26.517}\\
 \hline
 
  

    \end{tabular}%
}

  \caption{Comparison of the quantitative performance of the original and hidden message scenes. We also report the average rendering speed and the adaptability of SIBR viewer (official rendering engine of 3DGS). \textbf{Best} results are marked in \textbf{bold} and the second best results are marked with \underline{underline}.}
  \label{tab_main}
\end{table*}

\section{Experiment}

\subsection{Implementation Details}
Our code is based on 3DGS and we train models for $30000$ iterations on NVIDIA A800 GPU with the same optimizer and hyper-parameters as 3DGS. Unless otherwise specified, $\tau$ and $k$ in Eq. \ref{eq：op} and Eq. \ref{eq：SH} are set to 0.25 and 17. 
\subsection{Experimental Settings}
 \textbf{Datasets:} Same as GS-Hider \cite{zhang2024gshider}, we conduct experiments on 9 original scenes from the Mip-NeRF360 \cite{barron2022mipnerf360}, and the correspondence between the hidden and original scenes also follows GS-Hider.\\
 \textbf{Baselines:} we compare our method with existing
3DGS steganography method GS-Hider \cite{zhang2024gshider} and SecureGS \cite{zhang2025securegs}. 
Meanwhile, we feed the rendering results of the original 3DGS to a U-shaped decoder and constrain it to output hidden scenes according to \cite{li2023steganerf}, called 3DGS+StegaNeRF.\\
\textbf{Metrics:} We evaluate the
synthesized novel view in terms of Peak Singal-to-Noise Ratio (PSNR) and Structural Similarity Index Measure (SSIM). Meanwhile, we use rendering FPS to evaluate our efficiency.

\subsection{Evaluation}
Firstly, we compare splats in splats with baselines in terms of fidelity, efficiency, robustness, security, and usability.\\
\textbf{a) Fidelity:} We use Novel View Synthesis (NVS) results to evaluate scene fidelity. As shown in Tab.~\ref{tab_main}, our method achieves the best NVS quality on both original and hidden scenes. Fig.~\ref{fig_compare} shows more visual details which indicate that our splats in splats achieves the most appealing visualization results without the artifacts seen in SecureGS or the erroneous textures and colors observed in GS-Hider.\\
\textbf{b) Efficiency:} Tab.~\ref{tab_eff} demonstrates that we achieve SOTA rendering and training efficiency. Our method achieves 100 rendering FPS, which is 3x faster than the existing 3DGS steganography methods using neural networks, such as GS-Hider. Our training time is only half of the existing methods.
 \begin{table}[h]
\setlength{\tabcolsep}{1.1mm}
  \centering
    \resizebox{1.0\linewidth}{!}
    {

     \begin{tabular}{ccccp{0.05cm}cc}
\hline\thickhline
    & \multicolumn{3}{c}{Usability}  & &\multicolumn{2}{c}{Efficiency}\\
 \cline{2-4}  \cline{6-7}
\multirow{-2}{*}{Method}& Pipe.& Attr. & Viewer & & Train time& FPS\\
\hline

  \centering GS+StegaNeRF  &\XSolidBrush&\Checkmark&\XSolidBrush&&  97 Mins &  22  \\ 
  \centering  SecureGS &\XSolidBrush&\XSolidBrush&\XSolidBrush&&  71 Mins&  36  \\
  \centering GS-Hider   &\XSolidBrush&\XSolidBrush&\XSolidBrush&&  119 Mins&  44 \\
     \centering Ours &\Checkmark&\Checkmark&\Checkmark&&  47 Mins &  118 \\
     \hline
    \end{tabular}%
}
  \caption{Comparison of the usability and efficiency. Pipe. and Atrr. represent maintaining vanilla 3DGS's rendering pipeline and attributes. 
 }
  \label{tab_eff}
\end{table}%

\noindent\textbf{c) Usability:} Tab.~\ref{tab_eff} also demonstrates the adaptability of splats in splats and baselines to the rendering engine (SIBR Viewer) provided by 3DGS. Our method can be directly integrated into the rendering pipeline provided by 3DGS, offering a seamless user experience that existing approaches cannot achieve. We also demonstrate whether all methods maintain the vanilla 3DGS's rendering pipeline (Pipe. in short) and attributes (Attr. in short). Tab.~\ref{tab_eff} also indicates that our method is the first framework that embeds 3D content in 3DGS itself without modifying any attributes and pipelines.

\begin{table}[h]
  \centering

    \resizebox{1.0\linewidth}{!}
    {
    {
    
    \begin{tabular}{cccp{0.02cm}ccp{0.02cm}cc}
     \hline\thickhline
  Method&\multicolumn{2}{c}{SecureGS}& &\multicolumn{2}{c}{GS-Hider}& & \multicolumn{2}{c}{Ours}\\ 
\cline{2-3} \cline{5-6} \cline{8-9}
  Ratio & PSNR$\uparrow$ &  SSIM$\uparrow$& & PSNR$\uparrow$ & SSIM$\uparrow$ & &PSNR$\uparrow$ &  SSIM$\uparrow$\\
     \hline

     5\%&22.131&0.711&   &25.179& 0.780&& \textbf{26.519}&\textbf{0.797} \\
     10\%& 20.418 & 0.678&    &25.179& 0.780 &&
     \textbf{26.518}& \textbf{0.797}\\
    15\%& 19.158 & 0.645 &   &25.179 & 0.780&&\textbf{26.518}&\textbf{0.797}\\
    25\%& 16.960 &0.577 &  &25.167& 0.780 &&\textbf{26.517} &\textbf{0.797}\\
    \hline
    \end{tabular}%
}}
  \caption{Quantitative comparisons of sequential pruning. Our method achieves better robustness with lower metrics decline under sequential pruning.}

  \label{sequential}%
\end{table}%

\begin{table}[h]
  \centering
    \resizebox{1.0\linewidth}{!}
    {
    {
    
    \begin{tabular}{cccp{0.02cm}ccp{0.02cm}cc}
    \hline\thickhline
 Method&\multicolumn{2}{c}{SecureGS} &&\multicolumn{2}{c}{GS-Hider} && \multicolumn{2}{c}{Ours}\\ 
\cline{2-3} \cline{5-6} \cline{8-9}
  Ratio & PSNR$\uparrow$ &  SSIM$\uparrow$ & &PSNR$\uparrow$ & SSIM$\uparrow$ & &PSNR$\uparrow$ &  SSIM$\uparrow$ \\
     \hline

     5\%&22.920 &0.727 & &24.923&  0.774&& \textbf{26.415}& \textbf{0.794}\\
     10\%& 22.596 & 0.721 &   &24.673 &0.767&& \textbf{26.375}&\textbf{0.793}\\
    15\%&22.280 &0.713 &  & 24.371 &0.760& &\textbf{26.346}&\textbf{0.793}\\
    25\%&21.485 &0.695 &  &23.661 &0.741&&\textbf{26.320}&\textbf{0.792} \\
    \hline
    \end{tabular}%
}}
  \caption{ Quantitative comparisons of random pruning. Our method achieves better robustness compared to SecureGS and GS-Hider under random pruning.}

  \label{tab_random}%
\end{table}%
\noindent\textbf{d) Robustness:} 
To evaluate the robustness of our method, we degrade the Gaussian primitives using sequential pruning and random pruning following \cite{zhang2024gshider}. For sequential pruning, where primitives with lower opacity are deleted, our method exhibits exceptional stability: under 25\% sequential pruning, our PSNR decreases by merely 0.002. This significantly outperforms both GS-Hider (0.012 drop) and SecureGS (substantial 5.171 degradation). Similarly, for random pruning (25\% removal), our approach maintains strong robustness with only 0.09 PSNR reduction, whereas GS-Hider suffers a 1.260 decline and SecureGS shows even greater vulnerability with 1.435 PSNR loss. The comparative results in Tab.~\ref{sequential} and~\ref{tab_random} demonstrate that our framework achieves superior robustness against pruning attacks compared to existing SOTA methods.

\begin{figure}[h]
  \centering
    \includegraphics[width=0.95\linewidth]{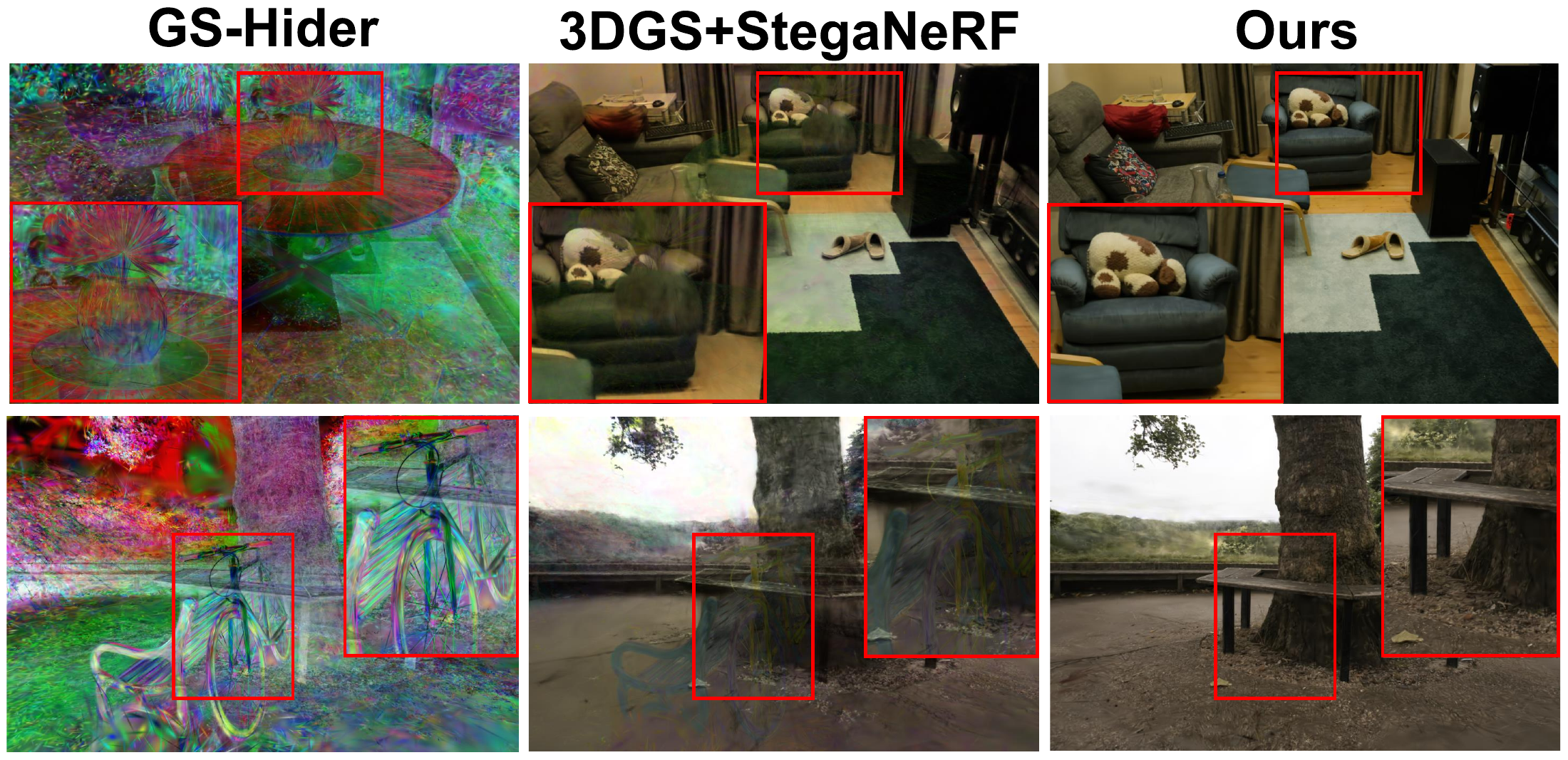}
    \caption{Comparison of rendering results. GS-hider and 3DGS+SteganeNeRF have obvious information leakage. \textbf{Better viewed on screen with zoom in}.}
    \label{fig_roc}
\end{figure}

\noindent\textbf{e) Security:} 
We analyze the security of splats in splats from the following aspects. From the view of point cloud file, it preserves the same structure as the original 3DGS. From the view of output, our scenes do not exhibit any artifacts or edges of hidden scenes and maintain high fidelity (shown in Fig.~\ref{fig_compare} and Tab.~\ref{tab_main}). Finally, we visualize the rendering results which can also be directly obtained by users.  Fig.~\ref{fig_roc} indicates that the rendering results of GS-Hider and 3DGS+Steganerf have obvious hidden information leakage while our method ensure the security.

\subsection{Ablation Study}

\noindent\textbf{a) SH/Opacity encryption:} We first conduct ablations to demonstrate the necessity of opacity mapping and SH encryption.  Quantitative results in Tab. \ref{tab_abop} demonstrate that opacity mapping offers a significant advantage in novel view synthesis for both original and hidden scenes, while SH encryption plays a crucial role in maintaining hidden scene fidelity. The aforementioned results are consistent with a fundamental principle that 3D assets are composed of two essential components \textbf{geometric structure} and \textbf{visual appearance.} Our method uses opacity mapping to hide geometric information and SH encryption to hide appearance.

\begin{table}[htb]
  \centering
    \setlength{\tabcolsep}{0.55mm}

    {
    \resizebox{\linewidth}{!}{
    \begin{tabular}{ccccp{0.2cm}ccc}
     \hline\thickhline
 \multirow{2}{*}{Method}&\multicolumn{3}{c}{Scene } & &\multicolumn{3}{c}{Message}\\ 
 \cline{2-4} \cline{6-8}
~&  $PSNR$ $\uparrow$&  $SSIM$  $\uparrow$& $LPIPS$  $\downarrow$& & $PSNR$ $\uparrow$&  $SSIM$  $\uparrow$& $LPIPS$  $\downarrow$\\
     \hline
     w/o op&24.209&   0.739& 0.301& &23.346& 0.767& 0.330\\
        w/o SH  &\textbf{26.795}&  \textbf{0.793}& \textbf{0.241}&& 11.092 & 0.432 & 0.331 \\
     SH+op (Ours)&26.749&  0.784& 0.262&& \textbf{26.517}&\textbf{0.797}& \textbf{0.308}\\

    \hline
    \end{tabular}%
}}
  \caption{  Quantitative results of ablations on SH/Opacity encryption. }

  \label{tab_abop}%
\end{table}%

\begin{figure}[h]
  \centering
    \includegraphics[width=\linewidth]{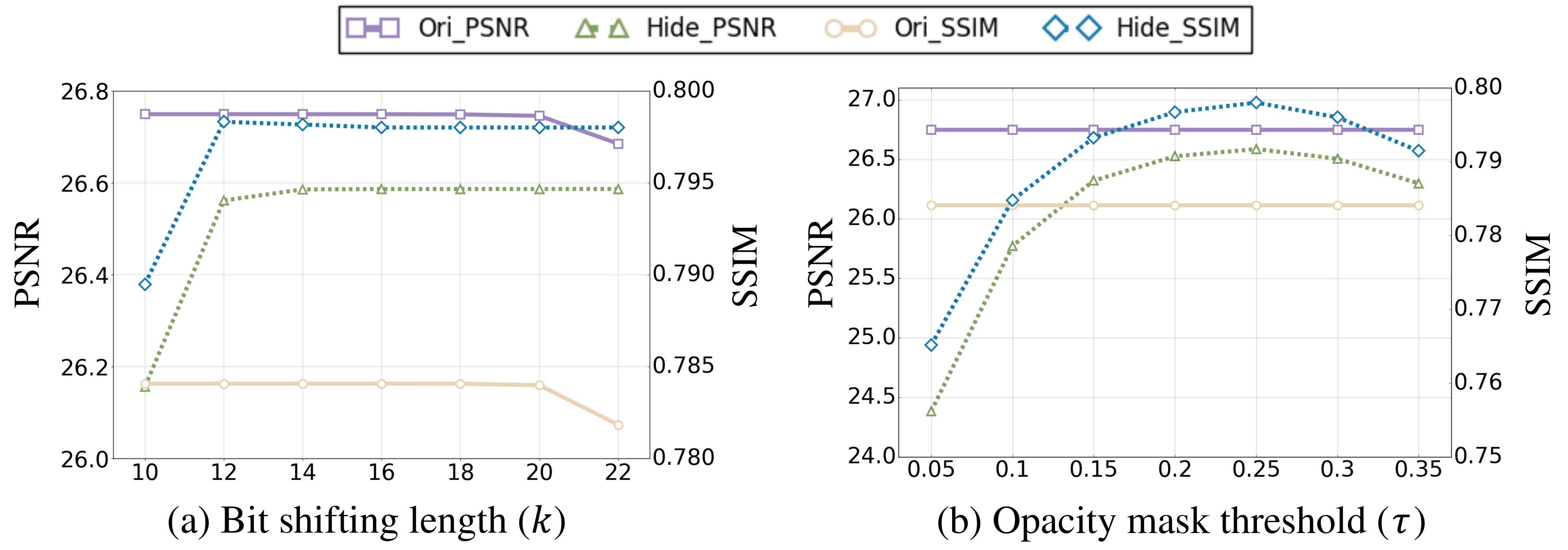}
    \caption{Effect of different shifting lengths $k$ for importance-graded bit encryption/decryption (sub-figure a) and thresholds $\tau$ for autoencoder-assisted opacity mapping (sub-figure b).}
    \label{fig_kbit_op}
\end{figure}

\begin{figure}[t]
  \centering
  \includegraphics[width=0.95\linewidth]{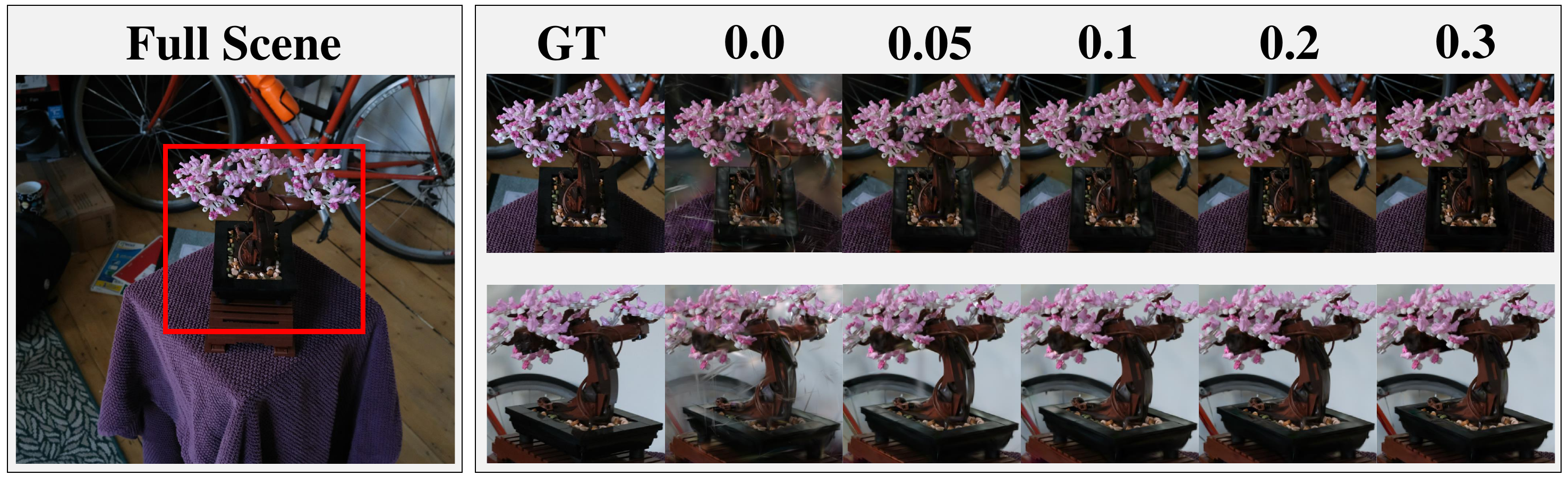}
  \caption{Visualization results of different opacity mask threshold. \textbf{Better viewed on screen with zoom in.} }
  \label{fig_opvis}
\end{figure}

\noindent\textbf{b) Opacity mask threshold:} We conduct ablations to investigate the effect of different opacity mask thresholds \(\tau\). We vary \(\tau\) in a wide range from 0.0 to 0.3 and plot the average PSNR and SSIM in Fig.~\ref{fig_kbit_op}. Results demonstrate that the opacity mask threshold has a significant influence on rendering quality and splats in splats achieve the best performance when opacity \(\tau\) is set to 0.25. The visualization results in Fig.~\ref{fig_opvis} further indicate that our opacity mask significantly reduces artifacts and geometric errors, leading to visually appealing rendering outcomes.

\begin{figure}[t]
  \centering
  \includegraphics[width=0.95\linewidth]{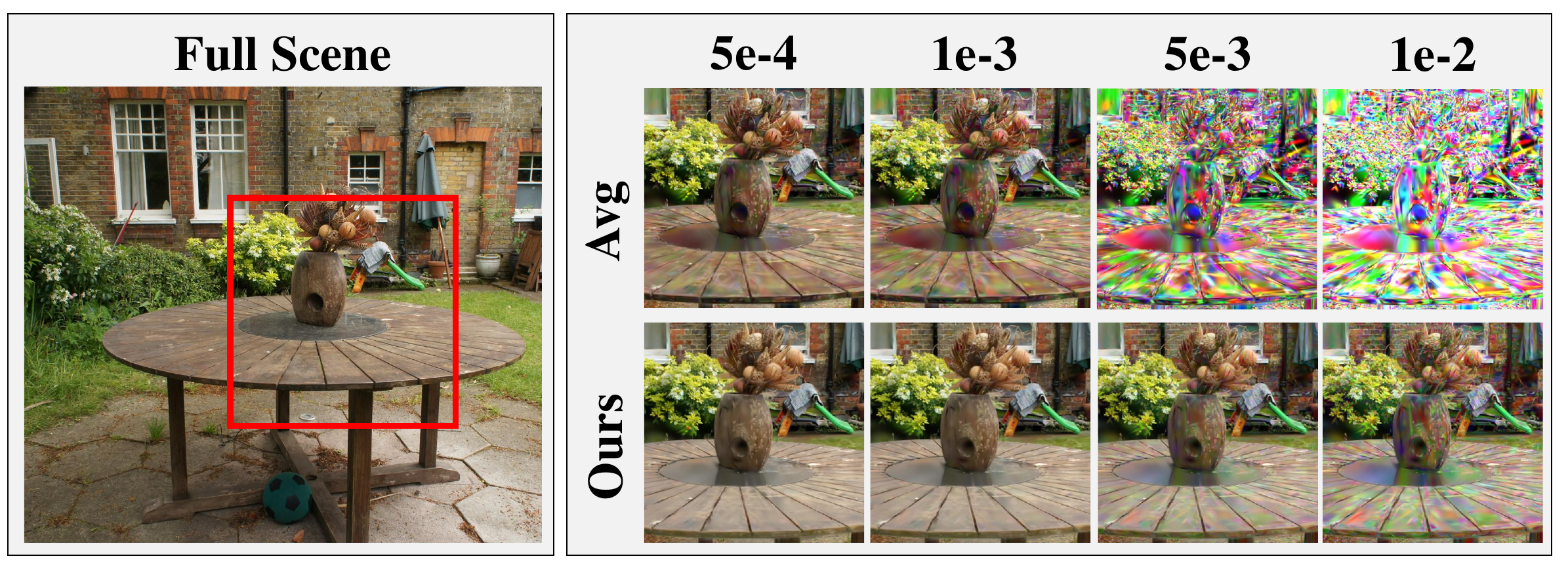}
  \caption{Visualization results under different noise levels. \textbf{Better viewed on screen with zoom in.}}
  \label{fig_noise}
\end{figure}

\noindent\textbf{c) Bit shifting length:} We vary the bit shifting length \(k\) from 10 to 22 and plot the average PSNR and SSIM. The results in Fig.~\ref{fig_kbit_op} demonstrate that \(k\) has minimal impact on rendering results of both original and hidden scene, which further demonstrates the robustness of our method.

\noindent\textbf{d) Importance-graded SH coefficient encryption:} We further conduct ablations to illustrate why importance-graded SH coefficient encryption is necessary. We compare the performance of our method with the average encryption (AVG) under different levels of Gaussian noise, with noise levels ranging from 0.0005 to 0.01. Quantitative results in Tab.~\ref{tab_noise} demonstrate that importance-graded encryption offers a significant advantage in noise resistance, which enhances both security and robustness. Fig.~\ref{fig_noise} confirms that as the noise level increases, AVG encryption causes a dramatic degradation in the quality of the recovered hidden scene. In contrast, splats in splats exhibits strong robustness against high-level noise, ensuring successful extraction of the hidden 3D information.

\begin{table}[h]
  \centering
    \setlength{\tabcolsep}{1.3mm}

    {
    
    \begin{tabular}{cccccc}
     \hline\thickhline
 \multirow{2}{*}{Method}&\multicolumn{4}{c}{Gaussian noise level }& \\
 \cline{2-5}
 &0.0005 &0.001&0.005&0.01& \multirow{-2}{*}{Average} \\ 

     \hline
     AVG&24.167 &21.991 &11.442& 7.471& 16.267 \\
     Ours&\textbf{24.577} &\textbf{24.509} &\textbf{22.797}& \textbf{20.032}& \textbf{22.979}\\
   \hline
    \end{tabular}%
}
  \caption{Quantitative results of PSNR for rendered hidden images under different noise levels.}
  \label{tab_noise}%
\end{table}



\section{Conclusion}
We propose splats in splats, an effective and flexible steganography framework for 3D Gaussian splatting. To the best of our knowledge, splats in splats is the first 3DGS steganography method that ensures security, fidelity, robustness, and rendering efficiency while maintaining usability and scalability. By carefully designed importance-graded SH coefficient encryption and autoencoder-assisted opacity mapping, splats in splats injects 3D information into the original 3D scenes presented by 3DGS while fully preserving the attributes of the vanilla 3DGS. Extensive experiments indicate that splats in splats achieves state-of-the-art rendering efficiency and fidelity while being well-suited for deployment across diverse applications. This paper offers a promising outlook on provenance verification in 3DGS and calls for more effort on user experience and scalability.
\textbf{Limitation:} Splats in splats has some impact on view-dependent details in both original and hidden scenes.  We will continue to improve our method to make it applicable to more scenes.


\section{Acknowledgments}
This work was supported by National Science and Technology Major Project (Grant No. 2022ZD0116305), National Nature Science Foundation of China under Grant No. 62088102, 62572314, 62202303, and 62471301, and the Beijing Natural Science Foundation (Grant Nos. F251020 and JQ24023).

\bibliography{aaai2026}

\end{document}